\documentclass[10pt,twocolumn,letterpaper]{article}

\usepackage[pagenumbers]{iccv} 

\usepackage{graphicx}
\usepackage{booktabs}
\usepackage{eccvabbrv}

\usepackage{comment}

\usepackage{amssymb}
\usepackage{derivative}

\graphicspath{{images/}}

\usepackage{acro}[=v2]
\acsetup{display-foreign = false}
\acsetup{foreign-format = {\scshape} }
\ExplSyntaxOn
\prop_new:N \l__acro_foreign_format_prop
\ExplSyntaxOff

\DeclareAcronym{NLLS}{short=NLLS, long=Non-Linear Least Squares}
\DeclareAcronym{IRLS}{short=IRLS, long=Iterated Reweighted Least Squares}
\DeclareAcronym{MAP}{short=MAP, long=Maximum a Posteriori}
\DeclareAcronym{MLE}{short=MLE, long=Maximum Likelihood Estimation}
\DeclareAcronym{FPGA}{short=FPGA, long=Field Programmable Gate Arrays}
\DeclareAcronym{PPA}{short=PPA, long=Pixel Processor Array}
\DeclareAcronym{ASIC}{short=ASIC, long=Application-specific integrated circuit}
\DeclareAcronym{SoC}{short=SoC, long=System-on-chip}
\DeclareAcronym{VO}{short=VO, long=Visual Odometry}
\DeclareAcronym{BA}{short=BA, long=Bundle Adjustment}
\DeclareAcronym{SLAM}{short=SLAM, long=Simultaneous Localization and Mapping}
\DeclareAcronym{VSLAM}{short=VSLAM, long=Visual Simultaneous Localization and Mapping}
\DeclareAcronym{DoF}{short=DoF, long=Degrees of Freedom}
\DeclareAcronym{IMU}{short=IMU, long=Inertial Measurement Unit}
\DeclareAcronym{ADMM}{short=ADMM, long=Alternating Direction Method of Multipliers}
\DeclareAcronym{BP}{short=BP, long=Belief Propagation}
\DeclareAcronym{GBP}{short=GBP, long=Gaussian Belief Propagation}

\newcommand{\multirefeq}[1]{%
  Equation\ifnum\@tempcnta>1 s\fi~\eqref{#1}%
}

\DeclareMathOperator*{\argmin}{arg\,min}

\def\plotwidth{0.84\linewidth}

\definecolor{iccvblue}{rgb}{0.21,0.49,0.74}
\usepackage[pagebackref,breaklinks,colorlinks,allcolors=iccvblue]{hyperref}

\def\paperID{11382} 
\def\confName{ICCV}
\def\confYear{2025}

\title{PixRO: Pixel-Distributed Direct Photometric Rotation Estimation\\ with Gaussian Belief Propagation}

\author{Ignacio Alzugaray$^{12}$ \quad Riku Murai$^{2}$ \quad Andrew Davison$^{12}$\\\\
{\normalsize ${}^{1}$ Dyson Robotics Lab \qquad ${}^{2}$ Department of Computing}\\
 {\normalsize Imperial College London, UK}\\
{\tt\small \{i.alzugaray,riku.murai15,a.davison\}@imperial.ac.uk}
}

\begin{document}
\maketitle
\begin{abstract}
Images are the standard input for most computer vision algorithms. However, their processing often reduces to parallelizable operations applied locally and independently to individual pixels. 
Yet, many of these low-level raw pixel readings only provide redundant or noisy information for specific high-level tasks, leading to inefficiencies in both energy consumption during their transmission off-sensor and computational resources in their subsequent processing.

As novel sensors featuring advanced in-pixel processing capabilities emerge, we envision a paradigm shift toward performing increasingly complex visual processing directly in-pixel, reducing computational overhead downstream.
We advocate for synthesizing high-level cues at the pixel level, enabling their off-sensor transmission to directly support downstream tasks more effectively than raw pixel readings.

This paper conceptualizes a novel photometric rotation estimation algorithm to be distributed at pixel level, where each pixel estimates the global motion of the camera by exchanging information with other pixels to achieve global consensus. 
We employ a probabilistic formulation and leverage \ac{GBP} for decentralized inference using messaging-passing.
The proposed proposed technique is evaluated on real-world public datasets and we offer a in-depth analysis of the practicality of applying \ac{GBP} to distributed rotation estimation at pixel level.

\end{abstract}
\section{Introduction}
\label{sec:introduction}
Visual sensors do more than merely capture light; they also pre-process data to enhance image quality using low-level algorithms such as gamma correction or noise suppression at the pixel or image level.
In the past decade, a new visual processing paradigm has emerged with the advent of smart sensors that shift part of the computational load toward on-sensor processing with more independent, computationally capable pixels. 
Notable examples include event cameras~\cite{lichtsteiner128x128120DB2008,poschQVGA143DB2011}, which asynchronously transmit data only from pixels detecting changes, and programmable sensor-processors  \cite{zarandy2011focal} such as the SCAMP sensor~\cite{carey100000Fps2013,boseSCAMP72023}, where each pixel (referred to as a pixel-processor) can locally execute a small program.
However, traditional vision algorithms still largely rely on raw pixel data transmitted off-sensor for batch processing as a globally accessible image in an external centralized processor (\eg, CPU, GPU).
As computation shifts toward the sensor, it becomes crucial to reconsider how algorithms can be deconstructed and distributed at the pixel level.

\begin{figure}[t]
    \centering
    \includegraphics[width=0.90\linewidth]{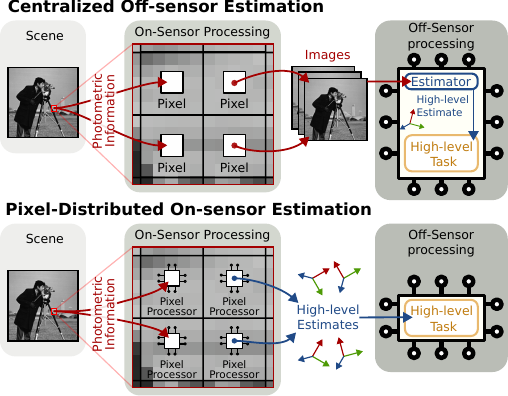}
    \caption{
    Most algorithms rely on receiving full-intensity images from traditional sensors, which are then processed off-sensor (\eg, by a CPU or GPU) to extract high-level cues for complex tasks (top). 
    Instead, we explore how to generate these high-level cues distributively at the pixel level, eliminating the need to transmit or process full images off-sensor and reducing computational overhead downstream while enhancing privacy (bottom).}
    \label{fig:variable}
\end{figure}

For high-level vision tasks such as \ac{VO}, only a small fraction of raw pixel readings actually provide non-redundant, highly informative data. Hence, transmitting full images off-sensor results in unnecessary energy costs and wasted computational resources processing uninformative pixels (\eg, textureless areas). 
This paper aims to explore how frame-to-frame direct photometric rotation estimation could be fully distributed at the pixel level, eliminating the need to transmit photometric information off-sensors (enhancing privacy) and without relying on close-to-sensor centralized processors (\eg, FPGAs, embedded GPUs). 
While our focus is mainly theoretical, we draw inspiration from currently available emerging sensor-processing architectures such as SCAMP, where each individual pixel have some limited computational capability and can communicate pixel-to-pixel.

We model our problem probabilistically as a factor graph and apply distributed inference using \ac{GBP}, a classical yet efficient and adaptable technique \cite{bishopPatternRecognitionMachine2006, davisonFutureMappingGaussianBelief2019} that has recently regained attention for distributed optimization \cite{ortizBundleAdjustmentGraph2020, sconaSceneFlowVisual2022, muraiRobotWebDistributed2022, nagataTangentiallyElongatedGaussian2023, muraiDistributedSimultaneousLocalisation2024}. 
In our approach, each pixel iteratively estimates the global camera motion in $\mathbb{SO}(3)$  via distributed consensus using only pixel-to-pixel message-passing. 
This paper is not limited to providing the theoretical grounds for such an algorithm, but also offers a comprehensive analysis of its characteristics and limitations. 

We believe this study pushes the envelope for future design of general-purpose pixel-distributed motion estimation algorithms aligned with the emerging sensor-processing paradigm and paving the way for a future where sensors directly synthesize and output high-level motion cues (\cref{fig:variable}). In summary, the contributions of this paper are:
\begin{itemize}
    \item We propose the first GBP-based direct-photometric frame-to-frame $\mathbb{SO}(3)$ rotation estimation algorithm to be distributed at the pixel level, relying solely on pixel-to-pixel message-passing.
    \item We provide a comprehensive description of the \ac{GBP} framework applied to 3D rotations, along with practical insights for real-world deployment. 
    \item We conduct extensive evaluations, comparing various communication strategies and variants against established baselines on public datasets, supported by an in-depth analysis of the method's characteristics.
\end{itemize}
\section{Related Work}
\label{sec:related_work}
As the field of \ac{VO} and \ac{VSLAM} reaches maturity \cite{davisonMonoSLAMRealTimeSingle2007, kleinParallelTrackingMapping2007,leuteneggerKeyframebasedVisualInertial2015, mur-artalORBSLAMVersatileAccurate2015,camposORBSLAM3AccurateOpenSource2021, engelLSDSLAMLargeScaleDirect2014,engelDirectSparseOdometry2018,qinVINSMonoRobustVersatile2018}, many established systems have been effectively migrated to hardware accelerators, offloading either the visual front-end \cite{weberrussFPGAAccelerationMultilevel2017, fangFPGAbasedORBFeature2017}, back-end  \cite{boikosSemidenseSLAMFPGA2016, qinPIBABundleAdjustment2019} or even the entire pipeline \cite{vemulapatiORBbasedSLAMAccelerator2022, asgariPISCESPowerAwareImplementation2020}.
More aligned with our paper, direct visual frame-to-frame rotation estimation is more closely related to optical flow estimation \cite{bakerDatabaseEvaluationMethodology2011} and direct image alignment \cite{lucasIterativeImageRegistration1981,bakerLucasKanade20Years2004}, which have also been extensively explored for hardware acceleration \cite{weiFPGAbasedRealtimeOptical2007,ishiiHighFrameRateOpticalFlow2012,bourniasFPGAAccelerationHorn2021}.
Most of these adaptations, however, largely parallelize the computation of resource-intensive subtasks within each method while the joint estimation still occurs in a centralized manner within a specialized hardware architecture.


Only a handful of papers, however, have addressed motion estimation using the near-sensor processing capabilities emerging in novel sensors such as   SCAMP~\cite{carey100000Fps2013,boseSCAMP72023}. 
For instance, in \cite{muraiBITVOVisualOdometry2020},  visual features are directly detected the sensor-processor and transmitted off-chip to perform \ac{VO} in a CPU.
In \cite{boseVisualOdometryPixel2017}, images are binary-thresholded into edges and aligned across frames using discrete in-plane image scaling, rotation and shifting (similar to \cite{alzugarayAsynchronousMultiHypothesisTracking2019,alzugarayHASTEMultiHypothesisAsynchronous2020}), to estimate  4-\ac{DoF} \ac{VO} within the sensor.
While this can only linearly approximate small discretized rotations for cameras with a large focal length \cite{bowen2022dimensions}, the approach is successfully extended to  6-\ac{DoF} \ac{VO}  in \cite{mcconvilleVisualOdometryUsing2020} by fusing information from an \ac{IMU}.
The work described in \cite{greatwoodPerspectiveCorrectingVisual2018} explores a similar image alignment technique to perform \ac{VO} against the ground-plane using near-sensor processing.
The design choices of these algorithms are, however, heavily tailored to the constraints imposed by the current SCAMP architecture (\eg, the lack of floating-point support or common operations such as multiplication and division). 
While we fully endorse the sensor-processing paradigm promoted by these approaches, we also envision pixel-processors in these sensors evolving toward more general computing capabilities, akin to those found in modern embedded CPUs. As such, our work focuses on the theoretical aspects of distributing rotation estimation at the pixel level rather than catering to the specific constraints of currently available architectures.


Distributed rotation estimation has been extensively studied for multiple images in the context of rotation averaging or \ac{BA} \cite{hartleyL1RotationAveraging2011, tronDistributedPoseAveraging2008,  erikssonConsensusBasedFrameworkDistributed2016, gaoIRADistributedIncremental2022,zhangDistributedVeryLarge2017}.
Nonetheless, these distributed approaches reason about relative rotations considering the images as an individual entity, often employing direct relative frame-to-frame measurements or groups of matched visual features.
In this paper, however, we further decompose the distributed estimation process to the most fundamental elements, \ie the pixels, and employ only direct pixel readings.

Among various distributed optimization frameworks \cite{halstedSurveyDistributedOptimization2021}, \ac{ADMM} \cite{boydDistributedOptimizationStatistical2011} has gained notable momentum and has been successfully applied to \ac{VO} and \ac{VSLAM} tasks \cite{erikssonConsensusBasedFrameworkDistributed2016,zhangDistributedVeryLarge2017,zieglerDistributedFormationEstimation2021}.
Nonetheless, traditional techniques such as \ac{GBP} \cite{bishopPatternRecognitionMachine2006,davisonFutureMappingGaussianBelief2019} have seen a reemergence due to their appealing probabilistic interpretation, asynchronicity and low-computational requirements.
\ac{GBP} has been successfully applied to distributed tasks such as multi-agent localization \cite{muraiRobotWebDistributed2022,muraiDistributedSimultaneousLocalisation2024}, \ac{VSLAM} \cite{hugHyperion2024}, or bundle-adjustment on a graph-processor \cite{ortizBundleAdjustmentGraph2020}. 
Closest to our work is \cite{sconaSceneFlowVisual2022} which performs pixel-wise \ac{VO} using \ac{GBP}, however, requires depth and random non-local connectivity.
\cite{nagataTangentiallyElongatedGaussian2023} uses \ac{GBP} and event camera for optical flow estimation, but unlike ours, cannot directly predict the motion of the camera.
In this paper, to the best of our knowledge, we explore for the first time the application of \ac{GBP} to frame-to-frame rotation estimation distributed at pixel level.

\section{Methodology}
\label{sec:methodology}

\newcommand{\ofp}[0]{\mathcal{T}}
\newcommand{\image}[0]{\mathcal{I}}
\newcommand{\diffimage}[0]{\mathcal{D}}
\newcommand{\rotation}{\mathrm{R}}
\newcommand{\rotstate}[0]{{R}(\variable)}
\newcommand{\motion}[0]{\mathcal{M}}
\newcommand{\disp}[0]{d}
\newcommand{\pixel}[0]{\mathbf{p}}
\newcommand{\point}[0]{\mathbf{P}}
\newcommand{\calib}[0]{\mathrm{K}}
\newcommand{\invcalib}[0]{\calib^{-1}}
\newcommand{\project}[0]{\pi}
\newcommand{\warp}[0]{\mathcal{W}}
\newcommand{\unwarp}[0]{\mathcal{W}^{-1}}
\newcommand{\unproject}[0]{\pi^{-1}}
\newcommand{\SOthree}[0]{\mathbb{SO}(3)}
\newcommand{\realset}[0]{\mathbb{R}}
\newcommand{\sothree}[0]{\mathfrak{so}(3)}
\newcommand{\homo}[1]{h(#1)}
\newcommand{\dehomo}[1]{h^{-1}(#1)}
\newcommand{\neigh}{\mathfrak{N}}
\newcommand{\Exp}{\text{Exp}}
\newcommand{\Log}{\text{Log}}
\newcommand{\meas}{\textbf{m}}
\newcommand{\measpred}{\hat{\textbf{m}}}
\newcommand{\factor}{f}
\newcommand{\factorset}{\mathcal{F}}

\newcommand{\obs}{\mathbf{z}}
\newcommand{\nobs}{\obs}
\newcommand{\hobs}{\hat{\mathbf{z}}}
\newcommand{\obsmodel}{h}
\newcommand{\obsdim}{{d_\mathbf{z}}}
\newcommand{\obsdimi}{{d_{\mathbf{z}_i}}}
\newcommand{\obsset}{\mathcal{Z}}
\newcommand{\obsnum}{{N_\obsset}}

\newcommand{\states}{X}
\newcommand{\stateset}{\mathcal{X}}

\newcommand{\variablenum}{{N_\variable}}
\newcommand{\gaussian}{\mathcal{N}}
\newcommand{\invgaussian}{\mathcal{N}^{-1}}

\newcommand{\graph}{\mathcal{G}}
\newcommand{\edge}{\mathbf{e}}
\newcommand{\edgeset}{\xi}
\newcommand{\energy}{E}
\newcommand{\msg}{\mathbf{m}}
\newcommand{\belief}{\mathbf{b}}
\newcommand{\veta}{\mathbf{\eta}}

\newcommand{\linstate}{\bar{\variable}}
\newcommand{\imageref}{\image_r}
\newcommand{\imagelive}{\image_l}

\newcommand{\variable}{\mathbf{v}}
\newcommand{\variableset}{\mathcal{V}}
\renewcommand{\prec}{\mathrm{\Lambda}}
\newcommand{\precinv}{\mathrm{\Lambda}^{-1}}
\newcommand{\info}{{\boldsymbol{\eta}}}
\newcommand{\transpose}{^{\mathrm {T}}}
\newcommand{\inv}{{^{-1}}}
\newcommand{\mean}{{\boldsymbol{\mu}}}
\newcommand{\cov}{\mathrm{\Sigma}}
\newcommand{\tangentplus}{\oplus}
\newcommand{\tangentminus}{\ominus}
\newcommand{\residualplus}{+}
\newcommand{\linpoint}{{\bar{\mean}}}
\newcommand{\linresidual}{\bar{\residual}}
\newcommand{\deltamean}{\boldsymbol{\tau}}
\newcommand{\jacobian}{\mathrm{J}}
\newcommand{\linjacobian}{\bar{\mathrm{J}}}
\newcommand{\optimal}{^{\mathbf{*}}}
\newcommand{\linfactor}{\bar{\factor}}
\newcommand{\gau}{\mathbf{g}}
\newcommand{\ggau}{\mathbf{G}}
\newcommand{\uporigin}{{{}^\origin}}
\newcommand{\uplinpoint}{{{}^\linpoint}}

\newcommand{\state}{\mean}
\newcommand{\residual}{\mathbf{r}}

\subsection{Gaussian Factor Graph}
\label{sec:factor_graph}
Gaussian Factor graphs \cite{dellaertFactorGraphsRobot2017}  
are bipartite graphs $\graph=\lbrace \variableset, \factorset \rbrace$ that model probabilistic relationships between Gaussian distributed variables $\variable \in \variableset$ and Gaussian distributed factors  $\factor\in\factorset$.
Gaussian distributions are characterized in standard form $\gaussian(\mean, \cov)$ with mean $\mean$ and covariance $\cov$, or in canonical form $\invgaussian(\info, \prec)$ with information vector $\info = \prec\inv\mean$ and precision matrix $\prec=\cov\inv$.
Individual factors $\factor_i$ model probabilistic constraints over a  subset of  variables $\variable_{\factor_i} = \neigh(\factor_i) \subseteq \variableset$ according to the factor potential $f_i(\variable_{\factor_i})\propto \exp(-\energy_{\factor_i}(\variable_{\factor_i}))$, modelled with an inducing energy term \cite{lecunTutorialEnergyBasedLearning2007} evaluated on the variables' mean $\mean_\variable$, i.e. $E(\variable) = \energy(\mean_\variable)$.
These factors essentially shape the likelihood of any provided configuration $\variableset$ as $ p(\variableset) = \prod_i \factor_i(\variable_{\factor_i})$ given a set of independent observations, enabling \ac{MLE} as in:
\begin{align}
    \variableset^*= \argmin_\variableset -\log p(\variableset) = \argmin_\variableset \sum_{\factor_i \in \factorset} \energy_{\factor_i} (\variable_{\factor_i}). \label{eq:mle}
\end{align}
The energy induced by each factor is characterized by a residual $\residual$ as
 $\energy (\mean) =\frac{1}{2} \| \residual(\mean) \|^2_{\prec_\residual}$, with $\| \residual \|^2_\prec = \residual\transpose \prec \residual $, that is zero-centered Gaussian-distributed  $\residual\sim \invgaussian(\mathbf{0}, \prec_\residual)$.

\subsection{Incremental \ac{GBP} for 3D rotations} 

\label{sec:gbp}

While \cref{eq:mle} is often addressed by techniques such as Gauss-Newton or gradient descent, \ac{GBP} offers a simple paradigm to \ac{MLE} by locally passing Gaussian messages among factors and variables to compute the marginals of the joint distribution.
These message-passing steps are efficient as they have closed-form expressions under Gaussian assumption and can be applied independently at each node and factor without explicit synchronization.
This makes \ac{GBP} an ideal candidate for efficient distributed inference while providing an explicit probabilistic interpretation of the problem.
Here we briefly summarize the key steps and refer the reader to previous work on the topic for further details \cite{pearlProbabilisticReasoningIntelligent1988,bishopPatternRecognitionMachine2006,davisonFutureMappingGaussianBelief2019,ortizVisualIntroductionGaussian2021,muraiRobotWebDistributed2022, muraiDistributedSimultaneousLocalisation2024}.

In particular, the scope of our paper is focused on distributed rotation estimation and thus all the variables $\variable\in\variableset$ in our graph represent rotations in $\SOthree$.
This generally establishes non-linear relationships on the factors operating on such variables with respect to their underlying minimal representation, imposing the need for finding the solution of \ac{MLE} problem incrementally. 
Following \cite{muraiRobotWebDistributed2022}, let $\linpoint\in \SOthree$ be the linearization point for a given rotation variable, \ie its mean. 
We define a random variable as:
\begin{equation}    
\variable = \linpoint \oplus \uplinpoint\xi \text{, where } \uplinpoint\xi \sim \gaussian(0, \cov_\variable) 
\label{eq:liegauss}
\end{equation}
By defining $\variable$ through the random variable $\uplinpoint\xi\in \sothree$, since Lie algebra is isomorphic to vector space, all the commonly used probabilistic methodologies can be reused \cite{solaMicroLieTheory2020,barfootStateEstimationRobotics2017}.
Notations and operations from~\cite{solaMicroLieTheory2020} are leveraged for the rest of the paper for $\SOthree$; for example, $\oplus$ applies exponential map and composition in one operation.

The residual of any factor involving such a variable can be approximated using Taylor expansion:
\begin{align}
    \residual(\mean) &= \residual(\linpoint \tangentplus \deltamean ) \approx \residual(\linpoint) \residualplus\left.\frac{D \residual}{D\mean}\right\rvert_{\mean=\linpoint}\deltamean \\&=  \residual(\linpoint) \residualplus\jacobian(\linpoint)\deltamean = \linresidual \residualplus\linjacobian\deltamean,
\end{align}
and each energy term can be approximated into an incremental form as:
\begin{align}
    \energy(\deltamean) = \frac{1}{2} \| \residual(\linpoint\tangentplus\deltamean) \|^2_{\prec_\residual} \approx \frac{1}{2} \|  \linresidual \residualplus\linjacobian\deltamean\|^2_{\prec_\residual}  \\ \propto \frac{1}{2}
    \deltamean\transpose \linjacobian\transpose \prec_\residual \linjacobian \deltamean + ( \linjacobian\transpose \prec_\residual\linresidual)\transpose\deltamean,
\end{align}
which, in turn, transforms the original factor potential $\factor_i$ for the residual $\residual_i$ into the incremental, linearized form ${}^\linpoint \factor_i \sim \invgaussian(\bar{\info}_{\factor_i}, \bar{\prec}_{\factor_i})$, with $\bar{\info}_{\factor_i} =  -\linjacobian_i\transpose \prec_{\residual_i}\linresidual_i$ and $\bar{\prec}_{\factor_i} =\linjacobian_i\transpose \prec_{\residual_i} \linjacobian_i $, around the linearization point $\linpoint$.
Adopting these incremental forms and $\SOthree$ for \ac{GBP}, we define the following operation for conciseness:
\begin{align}
    \ggau &= \linpoint \boxplus \uplinpoint\gau = \linpoint \boxplus  \gaussian(\uplinpoint\deltamean,{}^\linpoint\cov) \\&= (\linpoint \tangentplus \uplinpoint\deltamean) \oplus \gaussian\left(0, \jacobian_r(\uplinpoint\tau) {}^\linpoint\cov \jacobian_r\transpose(\uplinpoint\tau) \right),\label{eq:gauplus}\\
     \uplinpoint\gau &= \ggau \boxminus \linpoint  =  \left({}^{\ggau}\linpoint \oplus \gaussian(0,{}^\ggau\cov)\right) \boxminus  \linpoint  \\&= \gaussian( \boldsymbol{\theta} , \jacobian_r^{-1}(\boldsymbol{\theta}) {}^{\ggau}\cov \jacobian_r^{-\mathrm{T}}(\boldsymbol{\theta})), 
     \label{eq:gauminus}
\end{align}
with $\boldsymbol{\theta} = {}^{\ggau}\linpoint \tangentminus\linpoint$
and $\jacobian_r(\mathbf{x})$ the right-jacobian  $\SOthree$ \cite{solaMicroLieTheory2020}.

In the following lines, we describe the main \ac{GBP} steps adapted to this incremental $\SOthree$ estimation. Note that individual nodes and factors might operate on different linearization points, and thus their transformation is crucial in integrating the exchanged information via message-passing.
All messages are parameterized Gaussian as in \cref{eq:liegauss}, \ie a $\SOthree$ element with an uncertainty in its tangent space.

\textbf{Factor-to-Variable messages}: Each factor $\factor_j \in \factorset$  is evaluated at linearization point $\linpoint$ obtained from the means of the its connected variables $\variable_i \in \neigh(\factor_j) \subseteq \variableset$ to produce an incremental factor potential $\uplinpoint\factor_j$. 
This is combined with the incoming variable-to-factor messages $\msg_{\variable_k \rightarrow \factor_j}$ at their  linearization points $\linpoint_k$ to produce the local outgoing factor-to-variable messages:
\begin{align}
\msg_{\factor_j \rightarrow \variable_i} &=\linpoint_i \boxplus\left[  \sum_{\substack{\variable \in\\\neigh(\factor_j)/\variable_i}}  \uplinpoint\factor_j(\variable) \prod_{\substack{\variable_k \in\\ \neigh(\factor_k) / \variable_i}}\msg_{\variable_k \rightarrow \factor_j} \boxminus \linpoint_k\right]
.
\label{eq:gbp_f2v}
\end{align}
The outgoing message is computed by marginalization, then retraction onto $\SOthree$ around the outgoing variable's linearization point $\linpoint_i$.

\textbf{Variable-to-Factor messages}: each variable $\variable_i \in \variableset$ send message to each of factors it is connected to $\factor_j \in \neigh(\variable_i) \subseteq \factorset$ by leveraging its own mean as a linearization point $\linpoint_i$:
\begin{align}
\msg_{\variable_i \rightarrow \factor_j} = \linpoint_i \boxplus \left(\prod_{\factor \in \neigh(\variable_i) / \factor_j } \left(\msg_{\factor \rightarrow \variable_i}\boxminus \linpoint_i\right)\right)
~.
\label{eq:gbp_v2f}
\end{align}


\textbf{Variable belief update:}  Similarly, the marginal posteriors of each variable, \ie that best explain the constraints imposed by the factors, are computed by merging the incoming factor-to-variable messages in the local space:
\begin{align}
\variable_i =  \linpoint_i \boxplus \left(\prod_{\factor \in \neigh(\variable_i) } \left(\msg_{\factor \rightarrow \variable_i}\boxminus \linpoint_i\right)\right)\label{eq:gbp_belief}
\end{align}
Among these three steps, this last operation is the only one that modifies the linearization point of the problem by changing the $\variable_i$ as its mean is modified. 

Note that although \ac{GBP} lacks convergence guarantees for generic ``loopy'' graphs with cycles, it has been empirically demonstrated to produce compelling results for a variety of applications
\cite{ihlerLoopyBeliefPropagation2005,murphyLoopyBeliefPropagation2013,duConvergenceAnalysisBelief2018}. 

\subsection{Centralized Direct Rotation Estimation} 
\label{sec:centralized}
In general terms, our goal is to find the rotation, denoted by the $\state \in \SOthree$, that best describes the relative rotation between two image frames $\imageref$ and $\imagelive$ captured from the same static scene using a calibrated camera under pure-rotational motion.
These images are assumed to be distortion-free and grayscale.
We geometrically map individual pixel locations $\pixel$ from image $\imagelive$ onto image $\imageref$ using the warping function:
\begin{align}
    \warp(\pixel;\state) =  \project \left(\calib \state \invcalib\left[\pixel;1\right]^T\right),
\end{align}
where  
$\calib$ is the camera's intrinsic matrix,  and $\project(\point) = \left[p_x/p_z, p_y/p_z\right]^T$ is the projection function (explicitly indicated for completeness). 
We pose the problem as direct photometric image-alignment optimization between the warped image $\imagelive$ and image $\imageref$ over the whole image domain $\Omega$:
\begin{align}
        \state^* &= \argmin_\state \sum_{\pixel\in\Omega} \|\imagelive[\pixel] - \imageref[\warp(\pixel;\state)]\|^2 \label{eq:problem}
        ~.
\end{align}
This \ac{NLLS} problem is traditionally solved off-chip by transmitting images to a centralized CPU or GPU.
This method produces an optimal refinement of the current global estimate in each iteration by incorporating all available information from all pixels, which we refer to as \textit{centralized}.
When distributing the problem to individual pixels, updates rely solely on locally connected pixels, inevitably leading to suboptimal refinements.
Therefore, this original \textit{centralized} approach serves as the primary baseline, representing the best possible performance any distributed method can achieve.

Note that \cref{eq:problem} represents the simplest form of direct-image alignment and, thus, can only handle small camera rotations due to its small convergence basin, which will be reflected in our evaluations.
While more advanced techniques could be applied, such as multiscale pyramid \cite{bakerLucasKanade20Years2004}, we opt here to focus on this bare-bones version of the problem so that we can analyse better the characteristics of the proposed method without additional layers of complexity.

\subsection{Pixel-Distributed Direct Rotation Estimation}  
\label{sec:distributed}
In this section, we transform the centralized formulation described in \cref{eq:problem} into a pixel-level distributed problem for which we aim to estimate of the global rotation $\state_i$ at each single pixel $\pixel_i$, instead of a centralized estimate.
These per-pixel estimates $\state_i$  are the mean of the per-pixel variables $\variable_i$  (\cref{sec:factor_graph}), which we will iteratively solve for using \ac{GBP} (\cref{sec:gbp}).
Each of the per-pixel variables is fully constrained by factors described in the following lines (\cref{fig:variable}).

\textbf{Photometric Data Factor:}
This factor essentially mirrors the original formulation from \cref{eq:problem} at pixel level and indicates the photometric difference between individual pixel readings across the two considered images as in
\begin{align}
E_D^i (\state_i) = \tfrac{1}{2}\|\imagelive[\pixel_i] - \imageref[\warp(\pixel_i;\state_i)]\|^2_{\prec_D} \label{eq:factor_photo},
\end{align}
where each pixel is assumed to be aware of its location with respect to the camera center and the camera intrinsic $\calib$ so that $\warp$ can be applied.
In this paper, we assume that each pixel can have access to the photometric information of other pixels in the array (which could be implemented using routing/hopping mechanisms across neighbours).


It is crucial to note that, by breaking down the original problem into individual per-pixel photometric residuals applied on grayscale images, this factor only imposes a rank-deficient constraint onto the rotation estimate $\state_i$.
Consequently, other factors are required to fully constrain the problem, such as the prior and regularization terms.

\textbf{Prior Factor:}
This factor directly constraint the per-pixel estimated rotation through an educated guess $\hat{\state}_i$,
\begin{align}
E^{i}_P(\state_i) = \tfrac{1}{2}\|\state_i \ominus \hat{\state}_i\|^2_{\prec_P},
\label{eq:factor_prior}
\end{align}
where $\ominus$ represents distance metric of choice for comparing rotations  \cite{ parkDistanceMetricsRigidBody1995, parkSmoothInvariantInterpolation1997, huynhMetrics3DRotations2009}.
In our implementation, we choose the geodesic distance metric and use the previously known mean of the variable as the guess, \ie its linearization point $\hat{\state}_i=\linpoint_i$.
Thus, this factor only applies local constraint on the incremental step update of a variable, similar to a trust-region in traditional \ac{NLLS} for the \ac{GBP} framework.
While this factor might slow down the overall convergence of the system, it also stabilizes it in the presence of noise of highly irregular local gradients from \cref{eq:factor_photo}.


\textbf{Regularization Factor:}
In our algorithm, individual pixel variables exchange information with other  pixels to jointly reason about the global motion, which is represented by a regularization term:
\begin{align}
\energy^i_{R}(\state_i) = \sum_{\variable_j \in \neigh_R(\variable_i)} \tfrac{1}{2}\|\state_j  \ominus \state_i \|^2_{\prec_R} 
~,
\label{eq:factor_regularization}
\end{align}
where $\neigh_R(\variable_i)$ identifies the set of other neighbouring, connected pixel variables in the graph.
By enforcing locally connected pixels to yield the same global rotation (as they belong to the same sensor), this factor essentially propagates information across the pixels and achieves global consensus of the rotation estimate upon convergence.

While the photometric factor is the main driver for our rotational estimation, it tends to be extremely unreliable at the pixel level, either due to spurious noise, small convergence basins or simply because an individual pixel is observing an uninformative textureless area in the majority of the cases. 
In such situations, the regularization factor plays a fundamental role in the accurate estimation of the global rotation.
Yet a careful balance between photometric and regularization must be achieved to not completely disregard the image data while achieving a global consensus as we analyse in \cref{sec:evaluation:photovsregu}.
Given the importance of regularization in our approach, we discuss different topology patterns for connecting variables in the following section.


\subsection{Graph Topology}
\label{sec:topology}


\begin{figure}[t]
    \centering
        \includegraphics[width=\linewidth]{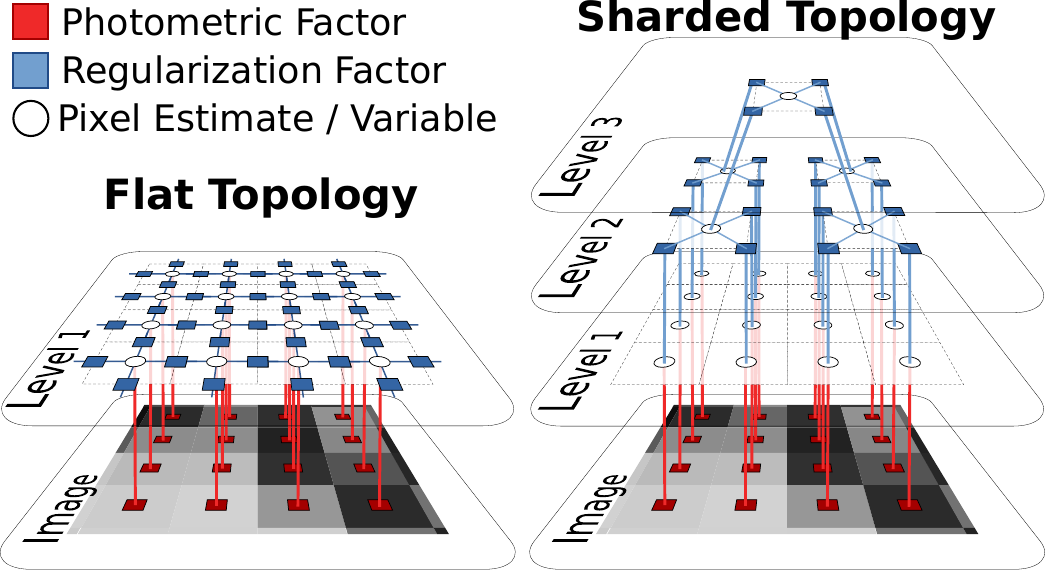}
    \caption{Pixel-processors arranged in a \textit{flat} (left) and \textit{sharded} (right) graph topologiesfor direct photometric frame-to-frame rotation estimation. 
    Each pixel-processor estimates a rotation variable (white) that can be constrained by photometric data factor (red) connected to the sensor image, and always includes prior factor (not depicted) and multiple regularization factors (blue) determined by the topology.}
    \label{fig:topology}
\end{figure}

Based on the presented framework, we explore two different configurations with distinctive graph topologies (see \cref{fig:topology}), leading to important performance differences as explored in \cref{sec:evaluation}.
Note that, in this paper, we are making use of a fixed, pre-determined graph topology with only local connectivity to simulate typical constraints that a pixel-processor currently exhibit (\eg, in SCAMP), whereas, in practice, the \ac{GBP} solver that acts upon this graph would allow us to mutate these connections on-the-fly, fostering future research.


\textbf{\textit{Flat} Topology:}
In this simple configuration, pixels are only connected to their immediate vertical and horizontal neighbouring pixels via regularization factors. 
Added to their photometric and prior factors, each pixel variable is connected to up to 6 factors in the graph.

\textbf{\textit{Sharded} Topology:} In this configuration, we consider two types of variables, photometric and sharded (level 1 and beyond level 1, respectively, in \cref{fig:topology}).
Photometric variables have access to raw pixel readings from the images and thus have associated photometric factors. 
Sharded variables are auxiliary variables that act as a bridge between other photometric and/or sharded variables but do not have associated photometric factors.
Instead, non-overlapping groups of $2\times2$ photometric variables are all connected to a single sharded variable in the next level via regularization factors.
In this new level of only sharded variables, non-overlapping groups of $2\times2$ are connected to the next level via regularization factors.
This process is repeated until a single sharded ``apex'' variable remains, at the top of the pyramid structure.
Both photometric and sharded variables make use of prior factors.
The motivation behind this topology is twofold: it propagates information more effectively between distant pixels and results in a tree-like graph without loops.
This enhances the convergence rate and accurately characterizes the system's uncertainty when using \ac{GBP} \cite{murphyLoopyBeliefPropagation2013}, in contrast to the \textit{flat} topology, as thoroughly analyzed in \cref{sec:evaluation}.
Additionally, raw image information never leaves the photometric variables in this configuration as sharded variables only leverages pure rotation estimates, which is not only ideal for privacy-preserving applications but also makes it interesting for the fusion of estimates from independent sensors by simply coupling their sharded variables with additional regularization factors, which we leave for future research.

\section{Implementation Details}

In contrast to most publicly available \ac{GBP} implementations in the literature \cite{ortizVisualIntroductionGaussian2021,davisonFutureMappingGaussianBelief2019}, here we do not explicitly consider three different steps of 
\cref{eq:gbp_v2f,eq:gbp_f2v,eq:gbp_belief} to be synchronously interleaved for factors and nodes.
Instead, we sequentially interleave only two steps: the updating of all of the factor-to-variable messages (\cref{eq:gbp_v2f}) and the update of all variable beliefs whereas variable-to-factor messages (\cref{eq:gbp_f2v}) are computed on-demand.
As both steps occur sequentially, we store the factor-to-variable messages in a local frame given a linearization point and also update them as soon as the linearization point changes, \ie just after a variable belief update, using \cref{eq:gauplus,eq:gauminus}.
We evaluate our algorithm using a dekstop machine and parallelize each of the \ac{GBP} steps using PyTorch (see Supplementary Material for additional details).

\renewcommand{\flat}{\textit{flat}\xspace}
\newcommand{\sharded}{\textit{sharded}\xspace}
\newcommand{\centralized}{\textit{centralized}\xspace}

\section{Experimental Evaluation}
\label{sec:evaluation}

\subsection{Data Generation}
For the proposed experiments, we generate our data from a public dataset of panoramic $360^{\circ}$ images, Pano3D GibsonV2 \cite{xiaGibsonEnvRealWorld2018, albanisPano3DHolisticBenchmark2021}, obtained from the projection of 3D scans in the real-world.
On each run, poses are generated by randomly selecting two rotations around the same camera center, as discussed in \cref{sec:centralized}, limited to rotations of $1^{\circ}$ between these frames. 
We synthesize the images from each pose by projecting equirectangular images with a distortion-free, projective camera model of  $60^{\circ}$ field-of-view and a resolution of $128$x$128$px.

Our evaluation mainly explores the characteristics of the proposed pixel-distributed direct rotation estimation method, emphasizing and analyzing the capabilities and limitations of the \flat and \sharded configurations (\cref{sec:topology}).
As a baseline, we consider a traditional \centralized configuration, where the NLLS problem in \cref{eq:problem} is optimized using a simple gradient descent.
Note that this \centralized configuration is expected to perform more effectively than the proposed distributed ones, as each optimization step has direct access to information from all the pixels to yield the best estimate.
As such, the \centralized baseline is only included for reference, whereas we focus our evaluation on the details of how to distribute the problem at pixel level.

In the following experiments, we report the normalized average rotational error as our main metric.
Unless otherwise specified, this is computed as the average geodesic distance between all individual rotation estimates at each variable and the ground-truth frame-to-frame global rotation, normalized with respect to the magnitude of such a rotation.
To overcome spurious effects, the reported metrics are collated from a collection of up to 50 different runs for each experiment, providing a more statistically meaningful result.
In our evaluation, we will considering modelling  factors from \cref{sec:distributed} with isotropic noise so that $\prec_i = \mathrm{\Sigma}_i\inv= \sigma_i^{-2}\mathrm{I}$, and specifying $\sigma_P$, $\sigma_D$ and $\sigma_R$, for  prior (\cref{eq:factor_prior}), photometric data  (\cref{eq:factor_photo}), and regularization (\cref{eq:factor_regularization}) terms.

\subsection{Performance of Pixel-Distributed and Centralized Rotation Estimation}

\begin{figure}[t]
    \centering
    \includegraphics[width=\plotwidth]{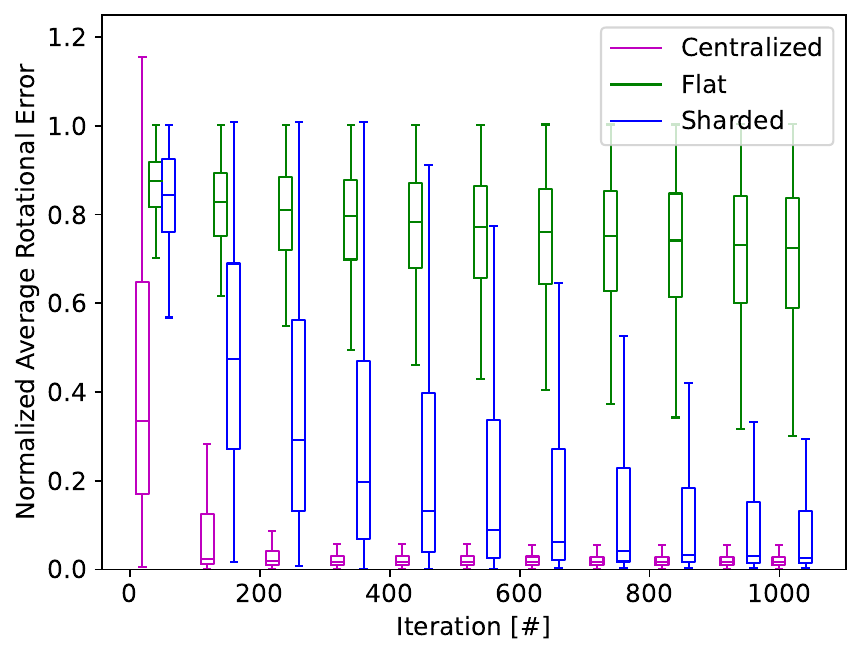}
    \caption{Progression of the error with respect to the number of optimization steps for the \centralized configurations or \ac{GBP} iterations for the distributed ones, \flat and \sharded.}
    \label{fig:convergence_average}
\end{figure}

In this experiment, we present a head-to-head comparison of all considered configurations: \flat, \sharded, and \centralized.
For fairness, we experimentally establish the best set of parameters for all configurations, employing $\lbrace\sigma_P,\sigma_D,\sigma_R\rbrace =\lbrace 10^{-2},10^{-1},10^{-2}\rbrace$ for the \flat topology and $\lbrace\sigma_P,\sigma_D,\sigma_R\rbrace =\lbrace 10^{-2},10^{-1},10^{-4}\rbrace$ for the \sharded topology.
Factor noise is often modelled according to experimental residual evaluation, which should be independent of the underlying topology.
However, in this case, the topologies present notably different behaviour for the same parameters (see \cref{sec:evaluation:photovsregu}).
The chosen parameters strike a good balance between accuracy, convergence and stability for the distributed configurations.

In \cref{fig:convergence_average}, we observe that the \sharded topology performs generally better than the \flat topology.
The former is able to more effectively converge and achieve rapid consensus of the global estimate.
Information from distant pixels can be propagated in fewer iterations via the hierarchical topology of \sharded variables in comparison to the \flat communication pattern.
Moreover, the \sharded topology offers a much better estimate of the global rotation at each variable in a reasonable number of iterations, performing on par with the \centralized upon convergence.
Conversely, the \flat topology often struggles to converge to a meaningful estimate, as consensus is usually not achieved.

As expected, the \centralized approach establishes the theoretical performance limit for both pixel-distributed approaches, as the information from all pixels is readily available at each iteration to yield the best estimate update. 
However, this is achievable only by globally aggregating the information from all the pixels at each iteration.
The pixel-distributed approaches explored here provide an alternative, parallelizable framework for motion estimation by further relaxing how to aggregate pixel information, potentially opening new research avenues at the intersection of hardware and software.


\subsection{Balance between Photometric Data Factor and Regularization Factors}
\label{sec:evaluation:photovsregu}

This section focuses on a comparison between the \flat 
and \sharded topologies, offering multiple insights by diving into the internals of the proposed framework.
This analysis is not only important to understand better the performance gap between the presented topologies, as evidenced in the previous experiment, but also identifies key aspects and potential improvements for other future approaches to be built within our estimation framework.
 
\begin{figure}
    \centering
    \begin{subfigure}{\plotwidth}
        \centering        
        \includegraphics[width=\textwidth]{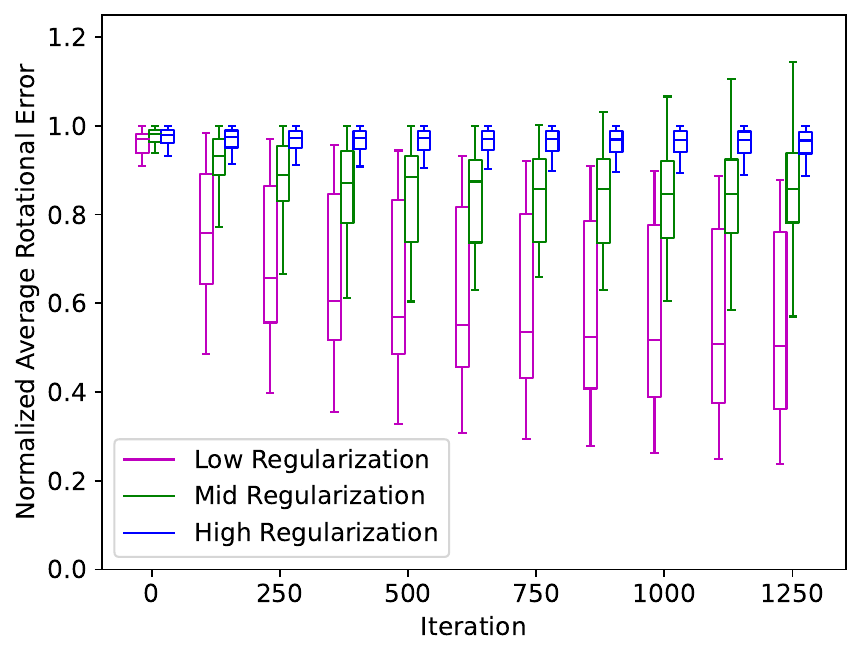}
        \caption{\textit{Flat} topology.}
    \label{fig:regularization_scan_flat}
    \end{subfigure}
    \begin{subfigure}{\plotwidth}
        \centering        
        \includegraphics[width=\textwidth]{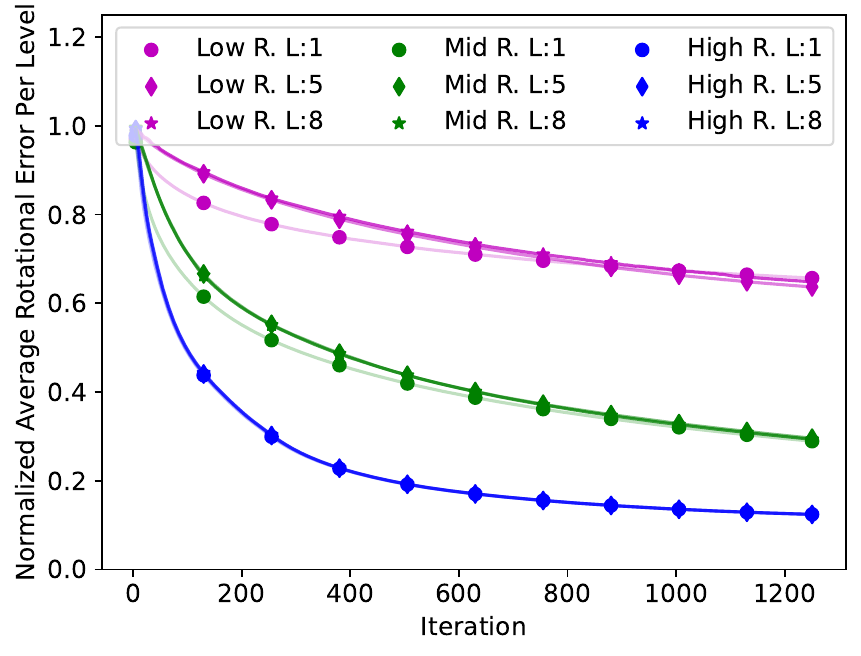}
        \caption{\textit{Sharded} Topology.}
        \label{fig:regularization_scan_sharded}
    \end{subfigure}
    \caption{Performance of the \flat and \sharded topologies with respect to the balance between photometric data and different strengths of regularization.}
    \label{fig:regularization_scan}
\end{figure}

As discussed in \cref{sec:distributed}, the per-pixel inference is mainly driven by the photometric data error.
Yet, regularization is necessary as it is the only means to communicate information across individual variables and achieve global consensus on the rotation estimates.
In this experiment we explore the fine balance between these two aspects by fixing  $\lbrace\sigma_P,\sigma_D\rbrace =\lbrace 10^{-2},10^{-1}\rbrace$ and varying the strength of the regularization:
 $\sigma_R =10^{-4}$ (High), $\sigma_R =10^{-3}$  (Mid), and $\sigma_R =10^{-2}$ (Low).
The results for both \flat and \sharded topologies are presented in \cref{fig:regularization_scan}.

Employing a \flat topology,  \cref{fig:regularization_scan_flat} illustrates that strong regularization halts the problem around the initial estimate, leading to non-meaningful estimates regardless of the number of iterations.
Alternatively, using weak regularization, some of individual variables yield a reasonable estimate of the global rotation.
However,  they effectively ignore the information from neighbouring variables and thus cannot achieve global consensus at convergence, leading to overall inaccurate estimation.
Interestingly, intermediate regularization strength can lead to instability of the system and, in some cases, divergence.
This is the result of poor variable estimates incorrectly receiving positive feedback from neighbours in the ``loopy'' graph; we will analyse this later.

\cref{fig:regularization_scan_sharded} offers an equivalent analysis for the \sharded topology.
Here we opt to leverage the hierarchical structure of this topology and report the rotational error summarized across individual levels, from the lowest with only photometric variables, to the top level, with only one apex variable (see \cref{sec:topology}).
For clarity, we show only three (L:1, L:5, L:8) of the eight levels,
and only show the means, as the statistical trends match those depicted in \cref{fig:convergence_average}.
Results indicate that rotation estimation accuracy at top levels consistently lags with respect to the variables closer to the photometric data in early iterations.
This is because higher-level variables mix conflicting estimates from different parts of the image, achieving consensus only after a few more iterations.
In the presence of strong regularization, however, this discrepancy is barely noticeable and all the individual estimates evolve jointly, making the system behave like a \centralized approach despite using only local message-passing.

\subsection{Stability Analysis on Graphs with Loops}
\label{sec:evaluation:stability}
\begin{figure}
    \centering
    \begin{subfigure}{\plotwidth}
        \centering        
        \includegraphics[width=\textwidth]{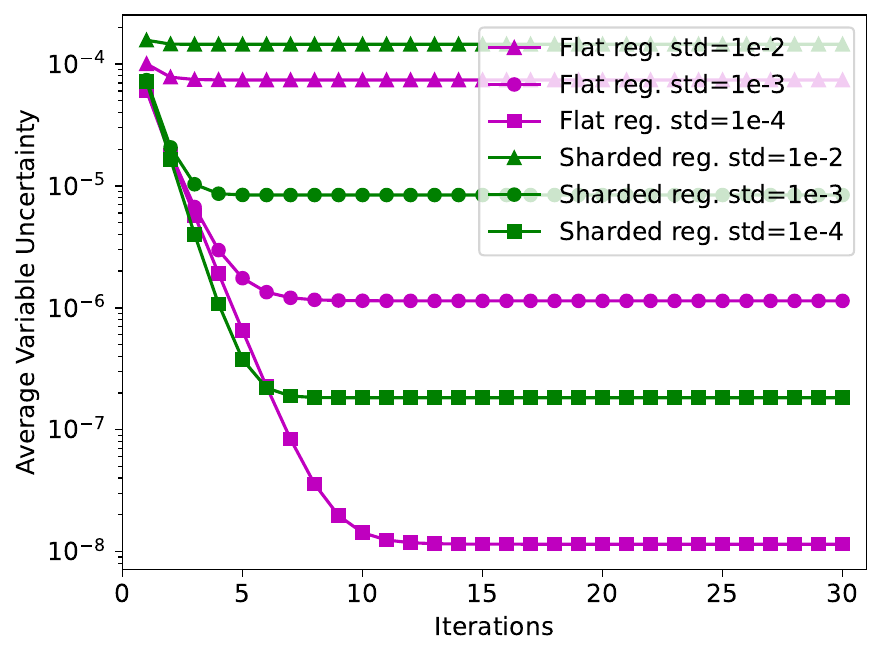}
        \caption{Uncertainty of \flat and \sharded topologies.}\label{fig:uncertainty_analysis}
    \end{subfigure}    
    \begin{subfigure}{\plotwidth}
        \centering        \includegraphics[width=\textwidth]{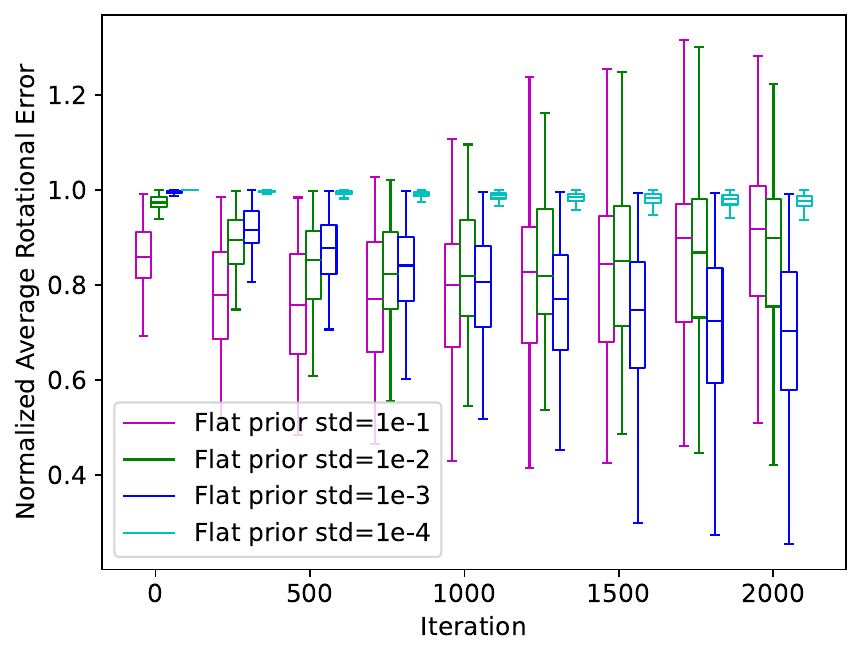}\label{fig:prior_strengthening}
        \caption{Prior strengthing for \flat topology.}
    \end{subfigure}
    \caption{Analysis of the overconfidence of \flat topology (top) and its effects on the stability of the system, that can be mitigated by varying prior strength (bottom).}
\end{figure}

Previous results indicate a noticeable difference in response to strong regularization between \flat and \sharded topologies.
In the former, estimation achieves consensus but to an incorrectly estimated global rotation near initialization, whereas the latter achieves its best-performing configuration.
In this section, we investigate the root cause for this difference, which can be explained by the fact that \flat topology creates a graph with multiple loops whereas the \sharded topology is essentially a tree.
Estimates from \ac{GBP} become more overconfident the ``loopier'' the factor graph is, which is a well-known effect described in the literature \cite{ihlerLoopyBeliefPropagation2005,murphyLoopyBeliefPropagation2013}.
To evidence this issue, \cref{fig:uncertainty_analysis} indicates the average variable uncertainty for each of the topologies, measured as the Frobenius norm of their covariance. 
In this analysis, we again explore three different degrees of regularization as in \cref{sec:evaluation:photovsregu}.
We observe that, for the same set of parameters, the \flat topology consistently yields overall lower uncertainty in the variables compared to the \sharded topology.
Note that this holds even when the \flat topology has fewer factors that constrain the graph when compared to the \sharded topology.

The overconfidence in variable estimates due to graph loops lead to positive feedback of poorly estimated variables, causing unstable behaviour as depicted in \cref{fig:regularization_scan_flat}.
While keeping the same set of parameters, $\lbrace\sigma_D,\sigma_R\rbrace =\lbrace 10^{-1}, 10^{-3}\rbrace$, we can mitigate the instability of the system by strengthening the prior $\sigma_P = \lbrace 10^{-1}, 10^{-2}, 10^{-3}, 10^{-4}\rbrace$ applied at each \ac{GBP} update, as represented in \cref{fig:prior_strengthening}, at the expenses of a lower convergence rate.
We identify this stabilizing effect related to the prior to be closely related to other techniques applied to \ac{GBP} such as messaging damping \cite{muraiRobotWebDistributed2022,muraiDistributedSimultaneousLocalisation2024} or diagonal loading 
\cite{johnsonFixingConvergenceGaussian2009}.

\subsection{Robustness to Noise}

\begin{figure}
    \centering
      \begin{subfigure}{\plotwidth}
        \centering        
        \includegraphics[width=\textwidth]{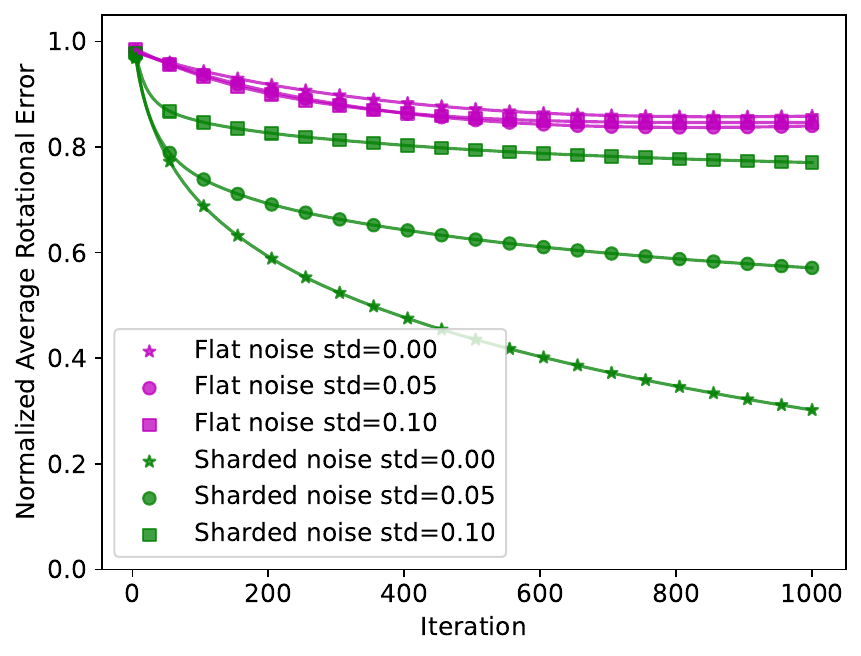}
        \caption{Weak regularization.}
        \label{fig:noise_weak_smoothness}
    \end{subfigure}
    \begin{subfigure}{\plotwidth}
        \centering        
        \includegraphics[width=\textwidth]{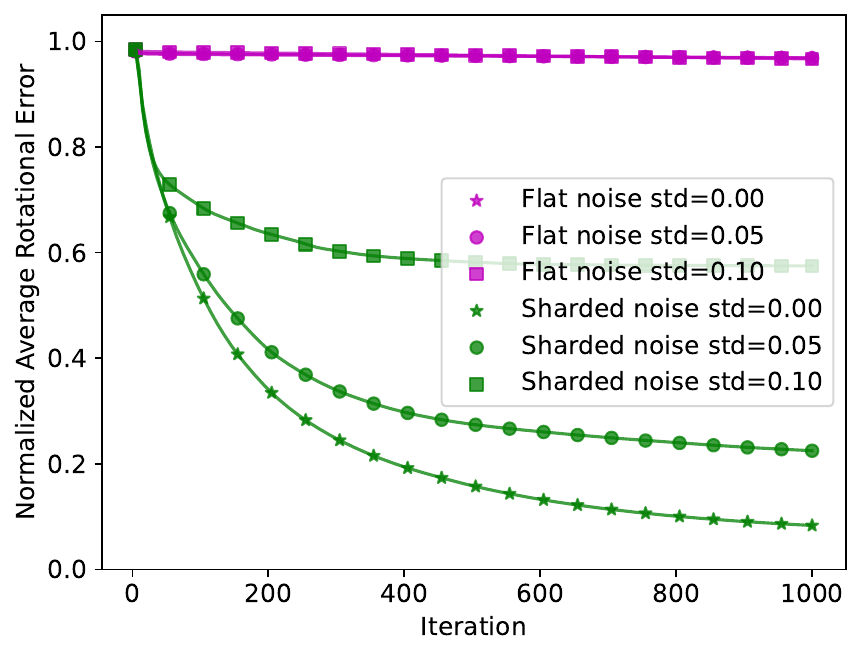}
        \caption{Strong regularization.}
        \label{fig:noise_strong_smoothness}
    \end{subfigure}    
    \caption{Performance of the \flat and \sharded configurations in the presence of noise employing weak or strong regularization.}
\end{figure}
The \flat and \sharded topologies also handle sensor noise differently.
Whereas in previous experiments we have used noise-free images, here we corrupt their values (normalized in $\left[0,1\right]$) with zero-mean Gaussian noise with standard deviation $\sigma_n = \lbrace 0.0, 5\cdot10^{-2},10^{-1}\rbrace$.
We consider two alternative configurations with either weak ($\sigma_R = 10^{-3}$) in \cref{fig:noise_weak_smoothness} or strong regularization ($\sigma_R = 10^{-4}$) in \cref{fig:noise_strong_smoothness}, leaving other parameters as in \cref{sec:evaluation:stability}.

While the performance of the system degrades as the level of image noise increases, we observe that the \sharded topology consistently outperforms the \flat one even in the presence of the strongest noise level.
Using a stronger regularization factor, however, helps to mitigate such detrimental effects while producing an overall better estimate as discussed in \cref{sec:evaluation:photovsregu}.
Despite the generally poor performance of the \flat topology, it is noteworthy that adding noise slightly improves the estimation capabilities of this topology.
This behaviour could be explained by interpreting the image noise as a way to disrupt the overconfidence of the synchronous \ac{GBP} steps, similar to improvement reported when randomly dropping messages \cite{muraiDistributedSimultaneousLocalisation2024}.

\section{Conclusions}

This paper introduces the first \ac{GBP}-based pixel-distributed frame-to-frame direct photometric $\SOthree$ rotation estimator that relies solely on local message-passing between variables.
We present a comprehensive evaluation that goes beyond accuracy reporting, offering a detailed analysis of the challenges and intricacies involved in distributing a global estimation problem at the pixel level using \ac{GBP}.
We believe this work paves the way for future research into the co-design of advanced algorithms and vision sensors with enhanced on-sensor pixel-processing capabilities.
The proposed pixel-distributed algorithm advocates for a promising paradigm shift, enabling vision-based pipelines to rely high-level motion estimates synthesized directly within the sensor.

{
    \small
    \bibliographystyle{ieeenat_fullname}
    \bibliography{bib}
}

\end{document}


\maketitle
\section{Factor graph complexity analysis in pixel-distributed approaches}
In decomposing the original \textit{centralized} formulation into a pixel-distributed one—either \textit{flat} or \textit{sharded}—additional factors and/or variables must be introduced to estimate the global camera rotation.
The resulting factor graph in our formulation is inherently tied to the native resolution of the image, as summarized in \cref{tab:graph_size} for each configuration.
In our approach, the graph topology remains fixed across iterations; however, we consider exploring dynamic topologies in future work (\eg, dynamically removing edges or disregarding pixel-to-pixel messages).

During each \ac{GBP} iteration, each pixel exchanges information with its connected pixels through their regularization factors.
While our current implementation synchronizes pixel updates and communication across the entire array, \ac{GBP} naturally supports random message-passing, which eliminates the need for global synchronization.
In some cases, such asynchronous updates have been reported to even improve convergence \cite{muraiDistributedSimultaneousLocalisation2024}, which we plan to investigate further.

It is important to note that our implementations of these topologies follow a deliberately simple design to evaluate performance without introducing additional complexities.
Many refinements could be made, such as replacing one out of every four photometric variables with a single sharded variable in the \textit{sharded} topology (rather than simply adding them in the current formulation) to maintain the same variable count as the \textit{flat} topology, or selectively updating data factors to only a subset of pixels (effectively subsampling the image).

\begin{table}
    \centering
    
\begin{tabular}{lcccc}
\toprule
\small{Topology} & \small{\# Variables} &  \multicolumn{3}{c}{\small{\# Factors}}  \\
 & & \small{Photo.} & \small{Prior} & \small{Regul.}\\
\midrule
\small{\textit{Centralized}}    & 1 & $\resolution^2$  & 0 & 0 \\
\small{\textit{Flat}}           & $\resolution^2$ & $\resolution^2$ &  $\resolution^2$   &  $2(\resolution^2-\resolution)$  \\
\small{\textit{Sharded}}        & $\frac{4^{\log{\resolution}}-1}{3}$& $\resolution^2$ &  $\frac{4^{\log{\resolution}}-1}{3}$ & $\frac{4^{\log{\resolution}}-1}{3} -1$ \\
\bottomrule
\end{tabular}
    \caption{Number of variables and photometric data, prior or regularization factors for an image of  $\resolution\times\resolution$ pixel resolution.  }
    \label{tab:graph_size}
\end{table}

\section{Computational performance on traditional centralized processors}
Although our contribution primarily focuses on establishing the theoretical foundation for distributing direct photometric rotation estimation at the pixel level, no existing sensor-processor device with in-pixel processing capabilities currently supports such operations.
For instance, the current SCAMP architecture lacks arithmetic operations such as multiplication or division, making it unsuitable for our approach.

\begin{figure}
    \centering
    \includegraphics[width=\linewidth]{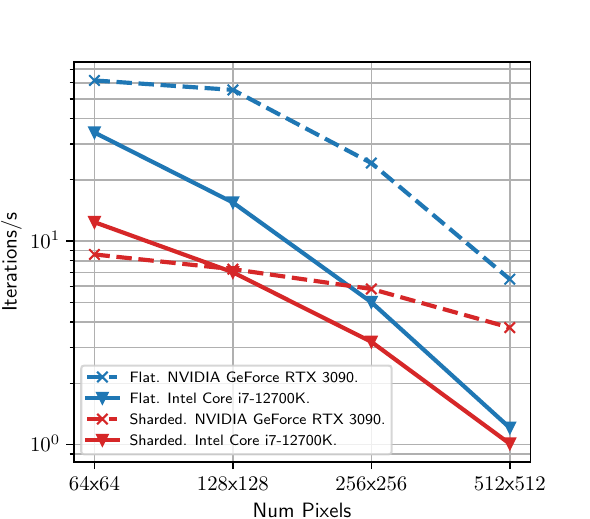}
    \caption{\ac{GBP} iteration rate as a function of the image resolution.}
    \label{fig:computation}
\end{figure}

As a result, we implement our algorithm using conventional, easily accessible consumer-grade centralized processors (\ie CPU and GPU) to evaluate its performance.
For each pair of images we must perform distributed inference on a mid- to large-scale graphs.
In a pixel-distributed paradigm, computational and memory loads would be naturally balanced across the sensor.
However, when executed on a centralized processor, the workload is concentrated creating a bottleneck.
\cref{fig:computation} reports the computational performance results from experiments conducted on a modern CPU and GPU.
It is worth noting that our implementation is largely unoptimized for these processors, as evidenced by the marginal performance gain of the GPU over the CPU, particularly at higher resolutions despite the high-degree of parallelism—indicating significant underutilization.
Crucially, while traditional hardware exhibits performance degradation as the number of pixels increases, a true pixel-processor-based implementation would scale naturally, distributing the computation and memory load as more pixels are added.

We anticipate that our algorithm would be best suited for emerging in-chip processing paradigms that diverge from traditional CPU/GPU architectures and instead leverage high-degrees of on-chip distributed parallelism, such as SCAMP-like architectures or other specialized graph processors \cite{ortizBundleAdjustmentGraph2020,jiaDissectingGraphcoreIPU2019}.

{
    \small
    \bibliographystyle{ieeenat_fullname}
    \bibliography{bib}
}